\documentclass{article}

\usepackage{PRIMEarxiv}

\usepackage[utf8]{inputenc} 
\usepackage[T1]{fontenc}    
\usepackage{hyperref}      
\usepackage{url}          
\usepackage{booktabs}      
\usepackage{amsfonts}      
\usepackage{nicefrac}       
\usepackage{microtype}     
\usepackage{lipsum}
\usepackage{fancyhdr}       
\usepackage{graphicx}       
\graphicspath{{media/}}     
\usepackage{amsthm}

\usepackage{amsmath}
\usepackage{amsthm}
\usepackage{booktabs}
\usepackage[switch]{lineno}
\usepackage{amssymb}
\usepackage{amsthm}
\usepackage{bm}
\usepackage{color}

\usepackage{graphicx}
\usepackage{multirow} 
\usepackage{rotating} 
\usepackage{xcolor}  
\usepackage{enumitem}

\usepackage{adjustbox}
\usepackage{threeparttable}

\usepackage[ruled,linesnumbered]{algorithm2e}

\title{A High-Performance Thermal Infrared Object Detection Framework with Centralized Regulation}

\author{
  Jinke Li \\
  Hangzhou Dianzi University \\
  \texttt{ljk221040091@163.com}\\
  \And
  Yue Wu\thanks{Correspondence to Yue Wu \texttt{yuewu@hdu.edu.cn}} \\
  Hangzhou Dianzi University \\
  \texttt{yuewu@hdu.edu.cn} \\
  \And
  Xiaoyan Yang\\
  Hangzhou Dianzi University\\
  \texttt{xiaoyanyang@hdu.edu.cn} \\
}

\begin{document}
\maketitle

\begin{abstract}
Thermal Infrared (TIR) technology involves the use of sensors to detect and measure infrared radiation emitted by objects, and it is widely utilized across a broad spectrum of applications.
The advancements in object detection methods utilizing TIR images have sparked significant research interest. However, most traditional methods lack the capability to effectively extract and fuse local-global information, which is crucial for TIR-domain feature attention.
In this study, we present a novel and efficient thermal infrared object detection framework, known as CRT-YOLO, that is based on centralized feature regulation, enabling the establishment of global-range interaction on TIR information.
Our proposed model integrates efficient multi-scale attention (EMA) modules, which adeptly capture long-range dependencies while incurring minimal computational overhead.
Additionally, it leverages the Centralized Feature Pyramid (CFP) network, which offers global regulation of TIR features.
Extensive experiments conducted on two benchmark datasets demonstrate that our CRT-YOLO model significantly outperforms conventional methods for TIR image object detection.
Furthermore, the ablation study provides compelling evidence of the effectiveness of our proposed modules, reinforcing the potential impact of our approach on advancing the field of thermal infrared object detection.
\end{abstract}

\keywords{Thermal Infrared Technology \and Object Detection \and Robust Representation Learning}

\section{Introduction}

Thermal Infrared (TIR) technology is a cutting-edge tool that harnesses the power of infrared radiation emitted by objects due to their temperature, providing valuable insights into temperature variations, heat distribution, and thermal anomalies.
By utilizing advanced sensors and cameras, TIR technology enables the visualization and measurement of thermal patterns, making it an indispensable asset across a diverse range of applications, including thermography \cite{gaussorgues2012infrared}, medical imaging \cite{lahiri2012medical}, environmental monitoring \cite{roberts2012synergies, li2019indoor, wang2023thermal}, security \cite{bhowmik2011thermal,metwaly2021fuzzy}, and surveillance \cite{wong2009effective,torresan2004advanced}.
Its versatile capabilities also extend to identifying overheating components, detecting insulation defects, monitoring thermal behavior in diverse environments, and aiding in medical diagnostics, highlighting its crucial role in various fields such as building diagnostics, industrial maintenance, and healthcare.
Moreover, TIR technology has found increasing integration into nighttime driving systems, offering a groundbreaking solution to enhance visibility and safety in low-light conditions.
By detecting and visualizing heat signatures from objects, TIR technology allows for the identification of pedestrians, animals, and other potential hazards that may not be visible with traditional headlights or vision systems, ultimately contributing to the reduction of accidents and the enhancement of road safety.
This innovative application demonstrates the invaluable potential of TIR technology in revolutionizing safety measures for nighttime driving, marking a significant advancement in automotive technology.

Conventional object detection methods normally use RGB images as the source input, where the benchmark datasets include Pascal VOC \cite{everingham2015pascal} and COCO \cite{caesar2018coco}.
Classical object detection models include Faster R-CNN \cite{ren2015faster}, SSD \cite{liu2016ssd} and YOLO models \cite{redmon2016you,redmon2017yolo9000,redmon2018yolov3,bochkovskiy2020yolov4,glenn_jocher_2021_5563715,chen2021you}.
However, RGB images have several disadvantages compared to TIR images:
Firstly, RGB images rely on visible light, making them less effective in low-light conditions, darkness, or environments with smoke, fog, or dust, where the visibility of objects is significantly reduced.
Secondly, RGB images are susceptible to variations in lighting conditions, making object detection and recognition challenging in environments with dynamic or harsh lighting.
Thirdly, they cannot detect objects based on their thermal signatures, limiting their effectiveness in scenarios where the detection of heat-emitting objects is crucial, such as in search and rescue operations or surveillance of thermal anomalies.
In contrast, TIR images capture the thermal radiation emitted by objects, enabling object detection even in low-visibility conditions.

Recently, research on TIR image object detection has attracted widespread interests \cite{liu2022cmx,chen2023igt,qingyun2022cross,wang2023tirdet,ding2023tir}.
TIR images offer several advantages over visible images due to their ability to capture and visualize heat radiation emitted by objects.
Some of the key advantages include:
(1) Temperature Sensing: TIR images provide temperature information about objects and the surrounding environment, allowing for the detection of heat signatures and variations. This is particularly useful in applications such as industrial monitoring, building diagnostics \cite{wuinvariant}, and biomedical thermography \cite{wang2018stimulation,wang2019ellagic,yang2020mechanism}.
(2) Low-Light and Nighttime Visibility: Unlike visible images, TIR images can capture thermal signatures even in low-light or nighttime conditions. This makes TIR technology invaluable for applications such as nighttime surveillance, search and rescue operations, and nighttime driving assistance systems.
(3) Penetration of Smoke, Dust, and Fog: TIR radiation can penetrate certain environmental obstructions such as smoke, dust, and fog, allowing for improved visibility and detection capabilities in challenging environmental conditions.


However, object detection in thermal infrared (TIR) images presents several significant challenges.
Firstly, the scarcity of large-scale annotated TIR image datasets hinders the training of deep learning-based object detectors, limiting the ability to harness the full potential of TIR technology for accurate and robust detection.
Secondly, the complex interplay between temperature differentials and object characteristics introduces additional intricacies into the detection process, requiring specialized techniques to effectively interpret thermal signatures and distinguish objects based on their thermal properties.
Moreover, the background scenarios in TIR image detection tasks are inherently complex, where environmental variations in thermal signatures can interfere with the accurate detection of objects, posing a significant challenge for reliable object detection in dynamic TIR environments.
Additionally, traditional object detection models designed for visible light-based images may not seamlessly adapt to TIR images due to the unique nature of thermal information and the distinctive features captured by TIR technology.
As a result, there is a pressing need for specialized methodologies and frameworks tailored specifically for TIR object detection to effectively address these challenges and maximize the potential of TIR imaging for diverse applications.

In this paper, we propose CRT-YOLO, a novel framework designed to specifically tackle the challenges inherent in object detection in TIR images, ultimately unleashing the full potential of TIR technology across diverse applications.
CRT-YOLO integrates efficient multi-scale attention (EMA) modules \cite{ouyang2023efficient}, meticulously engineered to adeptly capture long-range dependencies within TIR information while incurring minimal computational overhead.
This strategic integration empowers the framework to discern and prioritize critical long-range spatial relationships and contextual cues, enabling precise and comprehensive object detection capabilities even in complex TIR environments.
Meanwhile, the system incorporates the Centralized Feature Pyramid (CFP) network \cite{quan2023centralized}, which facilitates the global-range regulation of the TIR-domain features.
By synergistically leveraging centralized feature regulation and EMA modules, the CRT-YOLO framework not only excels in enhancing global-range interaction within TIR data but also demonstrates exceptional prowess in capturing nuanced long-range dependencies.
This approach establishes a robust and adaptive solution for accurate and efficient object detection in TIR images, surpassing the limitations of traditional detection models and paving the way for advanced thermal imaging applications across a spectrum of domains.
The experiments on two benchmark datasets, namely FLIR \cite{FLIR2019, zhang2020multispectral} and LLVIP \cite{jia2021llvip}, show that our proposed CRT-YOLO can efficiently improve the detection accuracy in the TIR domain.

The main contributions of this study are summarized below:
\begin{itemize}[left=0pt, label=•, itemsep=2pt, parsep=0pt, topsep=4pt]
  \item The study introduces the innovative CRT-YOLO framework tailored specifically for efficient and accurate object detection in Thermal Infrared (TIR) images.
  \item The CRT-YOLO framework integrates EMA modules to capture long-range dependencies within TIR information while minimizing computational overhead, addressing the unique challenges of TIR-based object detection.
  \item By leveraging centralized feature regulation of the CFP network, our CRT-YOLO framework demonstrates the capability to enhance global-range interaction within TIR data.
  \item Our extensive experiments on the benchmark datasets illustrate that it is capable of surpassing traditional detection models, which is particularly valuable in addressing the complexities of TIR environments.
\end{itemize}

\section{Related Works}

\subsection{YOLO models}

The YOLO (You Only Look Once) series of object detection models has significantly advanced the field of computer vision.
Through several iterations, each accompanied by improvements in model architecture and underlying technologies, the YOLO series has continuously pushed the boundaries of real-time object detection.

YOLOv1 \cite{redmon2016you}, the first model in the series, revolutionized object detection by treating it as a regression problem for spatially separated bounding boxes and class probabilities.
This innovative approach enabled real-time object detection by predicting bounding box coordinates and class probabilities without the need for complex post-processing steps.
Building upon the foundation of YOLOv1, YOLOv2 \cite{redmon2017yolo9000} introduced significant enhancements in both architecture and technologies, including batch normalization, high-resolution classifiers, anchor boxes for improved bounding box regression, and the use of multi-scale features for better detection performance.
Further refining the model architecture, YOLOv3 \cite{redmon2018yolov3} leveraged cutting-edge technologies and introduced pivotal features such as the use of a feature pyramid network (FPN) to capture multi-scale information, a prediction module enabling detection at three different scales, and the implementation of strong data augmentation techniques.
YOLOv3 also incorporated a more powerful backbone network, Darknet-53, for enhanced feature extraction and representation learning.
YOLOv4 \cite{bochkovskiy2020yolov4} consolidated advancements from previous iterations and introduced new features, leveraging architectural enhancements such as CSPDarknet53 as the backbone, spatial pyramid pooling (SPP) for improved feature representation, PANet for cross-scale feature fusion, and a modified YOLO head with advanced loss functions for enhanced training stability and convergence.
The YOLOv5 model \cite{glenn_jocher_2021_5563715}, designed to be flexible and adaptable, incorporates advanced training techniques such as mosaic data augmentation, label smoothing, and focal loss to enhance model generalization and robustness.
YOLOX \cite{yolox2021}, developed by MEGVII in 2021, represents a significant advancement in object detection by introducing a decoupled head for faster convergence and an anchor-free strategy to reduce model parameters and improve detection efficiency.
Later, YOLOv6 \cite{li2022yolov6} and YOLOv7 \cite{wang2023yolov7} were proposed subsequently, where model re-parametrization techniques and extended ELAN (E-ELAN) structures were introduced for real-time detection.
Now, the latest model is YOLOv8, which incorporates the benefits of the ELAN structure in YOLOv7, while incorprating the the Feature Pyramid Network (FPN) \cite{lin2017feature} and the Path Aggregation Network (PAN) \cite{li2018pyramid} structure to enhance the model’s feature fusion capability.

\subsection{Thermal Infrared Object Detection}
Object detection in thermal infrared (TIR) images has garnered significant attention due to the similarity in wavelength and its applications in various fields, such as autonomous driving.
While feature fusion in multispectral object detection has proven to enhance detection accuracy, it also introduces challenges related to memory storage and computational burden.
In the context of autonomous driving, memory usage and detection speed are crucial factors.
Therefore, the decision to focus on object detection in TIR images is both reasonable and advisable, given the importance of efficient memory utilization and swift detection processes.

Hwang \emph{et al.} \cite{hwang2015multispectral} constructed the first multispectral pedestrian benchmark and proposed a hand-crafted approach based on the aggregated channel feature, specifically tailored to extend infrared channel features, contributing to the advancement of pedestrian detection in multispectral images.
Later, Zhang \emph{et al.} \cite{zhang2020multispectral} developed a method for cyclically fusing and refining multispectral features with the aim of enhancing the consistency of both modalities, thereby improving the accuracy and robustness of multispectral object detection systems.
To address the misalignment problem between different modalities, Zhang \emph{et al.} \cite{zhang2019weakly} employed a region feature alignment module to anticipate feature offsets between RGB and thermal pictures, enabling more accurate and aligned feature representation across modalities.
Some researchers have delved into exploring the complementarity between visible and infrared images to leverage information from different modalities based on varying illumination conditions.
Guan \emph{et al.} \cite{guan2019fusion} proposed illumination-aware modules, allowing object detectors to dynamically adjust fusion weights based on predicted illumination conditions, leading to improved adaptability and robustness in varying lighting environments.
Subsequently, a memory-supervised residual module was introduced by Kim \emph{et al.} \cite{kim2022towards} to enhance the visual representation of single modality features by recalling multispectral modality features, effectively incorporating more discriminative features and improving the overall detection performance. Recent advances in robust representation learning have further inspired improvements in TIR detection frameworks \cite{lai2024fts, li2025causal}.
Due to the limitations of existing methods in modeling long-range dependencies across modalities, the Cross-Modal Fusion Transformer (CFT) was proposed \cite{qingyun2021cross}.
This method leverages local and global attention mechanisms to enhance the quality of feature fusion, allowing for more comprehensive and effective modeling of multispectral data for object detection.
The advancement of multispectral object detection, particularly in the context of pedestrian detection in TIR images, has been significantly driven by the development of innovative feature fusion techniques and the introduction of specialized modules to address specific challenges.

\section{YOLOv8 Baseline}
\begin{figure*}[!htbp]
	\centering
	\includegraphics[width=1\linewidth]{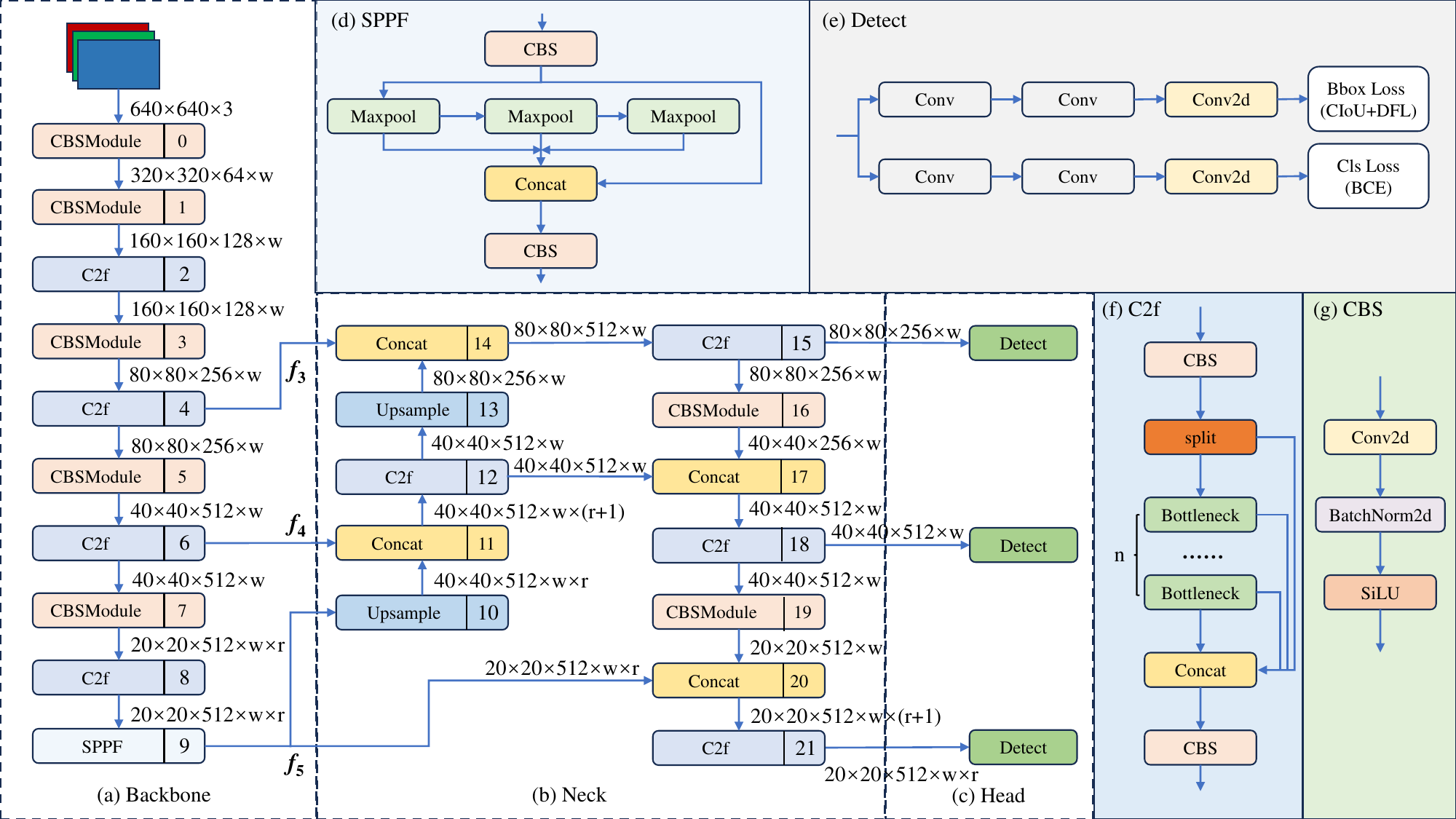}
	\caption{Illustration of YOLOv8 architecture.}
	\label{yolov8}
\end{figure*}  

Based on the progress of YOLO models, YOLOv8 \cite{Jocher_Ultralytics_YOLO_2023} is presented as an advanced and cutting-edge model that offers improved detection accuracy and speed.
The architecture of YOLOv8 is illustrated in Fig. \ref{yolov8}.
YOLOv8 utilizes CSPDarknet53 \cite{redmon2018yolov3} as the backbone network, which is a variant of the Darknet architecture used in YOLOv3.
The input features are down-sampled five times to obtain five different scale features.
This down-sampling is crucial for capturing features at multiple scales, which is essential for accurate object detection across different sizes and distances.
The model also incorporates the C2f module, which consists of a two-convolution cross-stage partial bottleneck. This module combines high-level features with contextual information to enhance detection accuracy.
By integrating contextual information, the model is able to better understand the relationships between different elements within the input data, leading to improved performance in object detection tasks.
Furthermore, the backbone network employs the spatial pyramid pooling fast (SPPF) module to pool the input feature maps to a fixed-size map for adaptive size output.
This is an improvement over the traditional spatial pyramid pooling (SPP) \cite{he2015spatial} approach, as SPPF reduces computational effort and latency by sequentially connecting three maximum pooling layers.
This allows for more efficient processing of the input feature maps, leading to improved speed and performance.

The neck of YOLOv8 incorporates FPN \cite{lin2017feature} and PAN \cite{li2018pyramid} structure to enhance the model's feature fusion capability.
This structure combines high-level and low-level feature maps using upsampling and downsampling techniques, facilitating the transfer of semantic and localization features.
The use of FPN and PAN enables the model to effectively fuse features of different scales, leading to a more comprehensive representation of the input data.

In addition to the FPN and PAN integration, YOLOv8 also employs an anchor-free model with decoupled heads to independently handle object detection, classification, and regression tasks.
This approach allows for more flexibility in handling different aspects of the detection process and can lead to improved overall performance.
The loss calculation process in YOLOv8 involves the use of the TaskAlignedAssigner \cite{feng2021tood} to determine positive and negative sample assignments.
The classification branch utilizes Binary Cross-Entropy (BCE) loss, while the regression branch employs the distribution focal loss and CIoU (Complete Intersection over Union) loss functions.
By using a combination of these loss functions, the model can effectively optimize both the classification and regression tasks, leading to improved accuracy in object detection.
YOLOv8 employs three sets of detectors, each with a different scale, to aid the model in recognizing objects of various sizes.
This multi-scale approach allows the model to effectively detect objects at different distances and sizes, leading to improved overall performance in object detection tasks.

\section{CRT-YOLO}

\subsection{Overview}
\begin{figure*}[!htbp]
  \centering
  \includegraphics[width=1\linewidth]{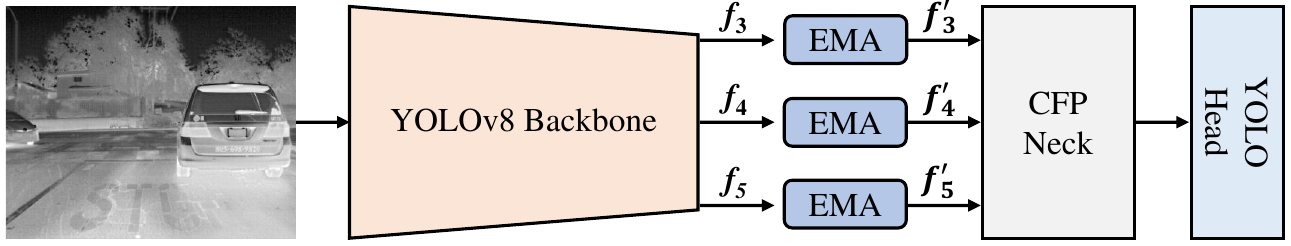}
  \caption{Illustration of the pipeline of our proposed CRT-YOLO.}
  \label{model}
\end{figure*}

The proposed CRT-YOLO pipeline, illustrated in Figure \ref{model}, commences with the thermal infrared (TIR) image as the primary input, which undergoes initial processing through the YOLOv8 backbone.
This yields the backbone model's outputs, designated as $f_3$, $f_4$, and $f_5$.
These output features are then subjected to further refinement via the application of Efficient Multi-Scale Attention (EMA) modules, resulting in the generation of enhanced feature representations denoted as $f_3^{'}$, $f_4^{'}$, and $f_5^{'}$.
Subsequently, these refined features are channeled into the Centralized Feature Pyramid (CFP) neck, where a combination of long-range and local interactions are effectively established.
This process allows for the fusion of multi-scale information, facilitating a more comprehensive understanding of the input TIR image.
The CFP neck plays a crucial role in integrating and consolidating the refined features to enable the subsequent stages of the pipeline to effectively utilize the enriched representations for accurate object detection and localization.
Additionally, the CFP neck ensures that the refined features maintain their spatial relationships, thereby preserving crucial spatial information during the feature fusion process.

\subsection{EMA module}
\begin{figure*}[!htbp]
  \centering
  \includegraphics[width=1\linewidth]{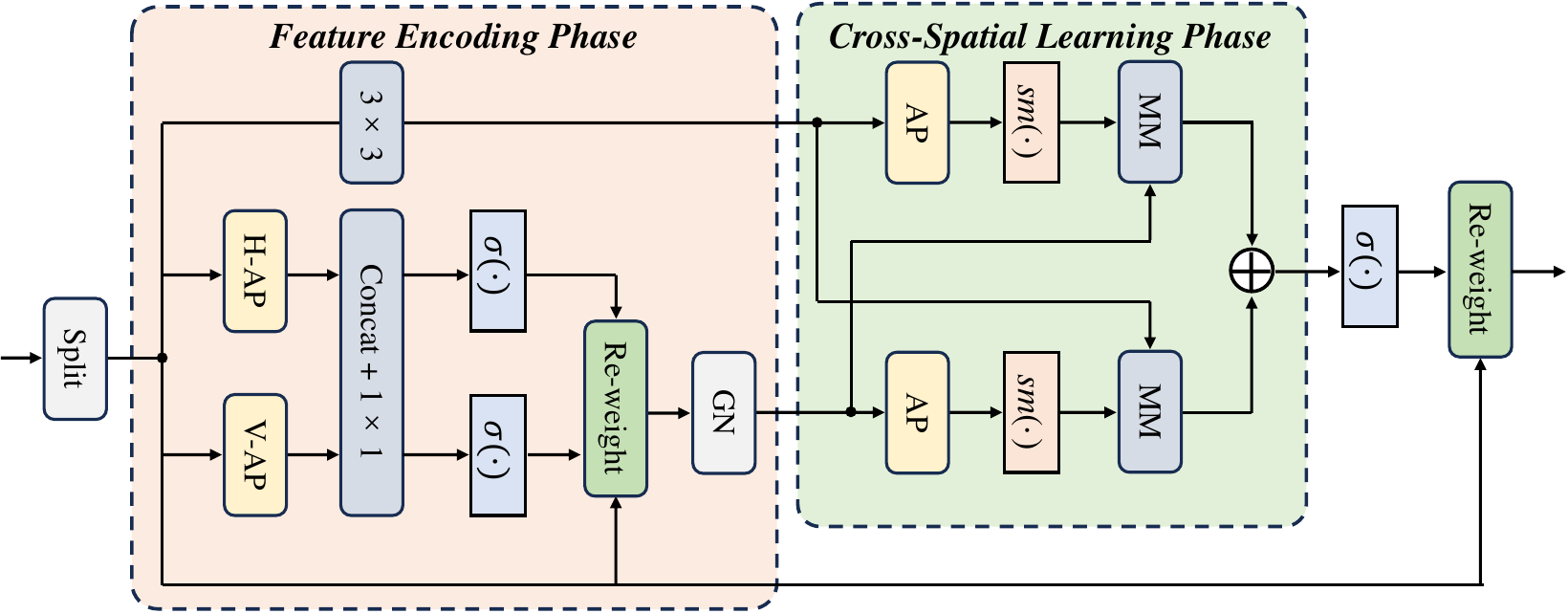}
  \caption{Illustration of EMA module. AP: Average Pooling; H-AP: Horizontal Average Pooling; V-AP: Vertical Average Pooling; GN: Group Normalization; $\sigma$($\cdot$): sigmoid function; sm($\cdot$): softmax function; MM: Matrix Multiplication.}
  \label{ema}
\end{figure*}

The Efficient Multi-Scale Attention (EMA) module, which is proposed by Ouyang \emph{et al.}, operates through a meticulously structured computational workflow that facilitates the enhancement of feature representations.
The workflow is illustrated in Fig. \ref{ema}.
EMA module commences with the input feature map being divided into G sub-features across the channel dimensions, where each sub-feature belongs to a specific group.
These sub-features are then subjected to parallel routes for feature encoding, global information encoding, and cross-spatial information aggregation.

In the feature encoding routes, the EMA module utilizes both $1\times 1$ and $3\times 3$ convolutions to capture long-range dependencies and multi-scale feature representations.
The $1\times 1$ branches also employ global average pooling operations to encode channel attention along the spatial directions, effectively capturing long-range dependencies and preserving precise positional information.
Concurrently, the $3\times 3$ convolutional branch facilitates the aggregation of spatial attention weight values to encode global spatial information.
This comprehensive architecture enables the module to recalibrate channel-wise relationships and produce pixel-level attention for high-level feature maps.
Subsequently, the outputs of the $1\times 1$ convolution are factorized into two 1D vectors, followed by the application of non-linear Sigmoid functions to model a 2D binomial distribution.
The resulting channel attention maps within each group are then aggregated through a simple multiplication, adapting to recalibrate the channel-wise relationships to achieve diverse cross-channel interactive features between the parallel routes in the $1\times 1$ branch.
By preserving precise positional information and recalibrating the raw input features, the model learns low-level detailed feature representations. Additionally, the raw input features undergo further processing through a $3\times 3$ convolution to expand the feature space.
The global average pooling along the horizontal and vertical dimensions can be formulated as:
\begin{equation}
  z_c^H(H)=\frac{1}{W} \sum_{0 \leq i \leq W} x_c(H, i)
\end{equation}
\begin{equation}
  z_c^W(W)=\frac{1}{H} \sum_{0 \leq j \leq H} x_c(j, W)
\end{equation}
where $H$ and $W$ denote the spatial size of the input feature $x_c$.

Followed by the parallel feature encoding phase, the cross-spatial learning phase is employed to extract the global-context information.
The cross-spatial aggregation uses two branches, one for $1\times 1$ convolution and the other for $3\times 3$ convolution.
We employ 2D global average pooling to encode global spatial information in both branches.
The outputs from each branch undergo transformations before being jointly activated for channel features.
By applying matrix dot-product operations and non-linear Softmax functions, we derive spatial attention maps capturing diverse scale spatial information.
These maps preserve precise spatial positional information and are utilized to aggregate output feature maps within each group.
The result feature efficiently integrates into modern architectures, emphasizing the representation of interests by modeling long-range dependencies and embedding precise positional information.
The 2D global pooling operation in this phase can be formulated as:
\begin{equation}
  z_c =\frac{1}{H\times W} \sum_{j}^{H}  \sum_{i}^{W} x_c(j, i)
\end{equation}

From a computational perspective, the EMA module's parallel routes for feature encoding and attention weight aggregation entail a series of matrix operations, including convolutions, global average pooling, and channel-wise recalibration.
These operations are performed in parallel across the sub-features, optimizing computational efficiency while maintaining a high degree of precision.
The resulting output of the EMA module is a feature map that emphasizes representations of interest, effectively embedding both long-range dependencies and precise positional information.
The advantages of the EMA module are rooted in its ability to learn effective channel descriptions without resorting to channel dimensionality reduction, efficiently model long-range dependencies, and embed precise positional information.
This comprehensive computational workflow, characterized by parallel feature encoding routes and attention weight aggregation, contributes to the substantial improvement of feature representations in computer vision tasks, thereby positioning the EMA module as a sophisticated and valuable tool in the realm of feature extraction and computer vision.

\subsection{CFP Network}
\begin{figure*}[!htbp]
  \centering
  \includegraphics[width=1\linewidth]{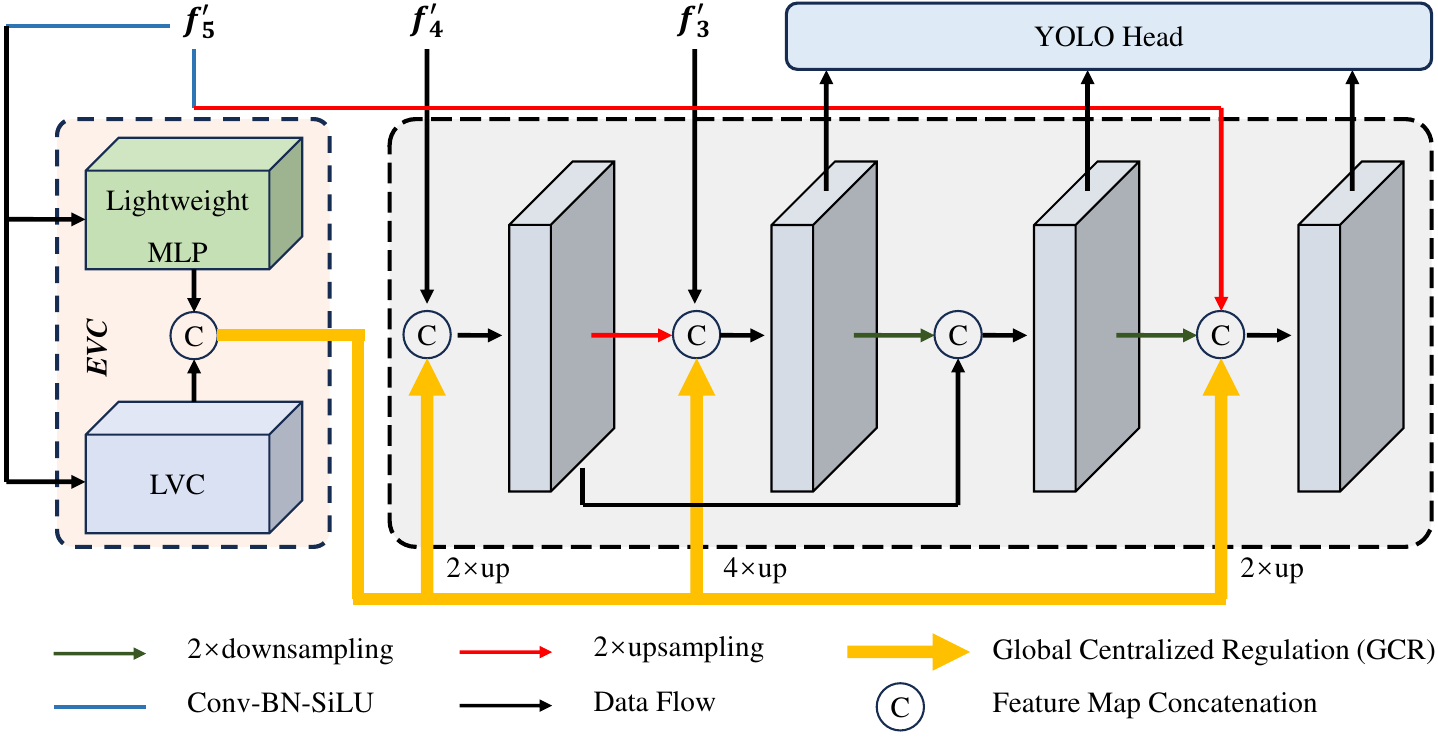}
  \caption{Illustration of Centralized Feature Pyramid (CFP) network.}
  \label{CFP}
\end{figure*}
The Centralized Feature Pyramid (CFP) network introduces a globally explicit centralized feature regulation approach, capturing both long-range dependencies and local corner regions for dense prediction tasks.
It efficiently obtains an all-round yet discriminative feature representation, leading to consistent performance gains in object detection tasks when integrated with state-of-the-art detection models.
The workflow involves a spatial explicit visual center scheme and a parallel learnable visual center mechanism to regulate the feature pyramid in a top-down fashion, utilizing explicit visual center information from the deepest intra-layer feature to regulate frontal shallow features.

We begin by inputting an arbitrary RGB image into the backbone network to extract a five-level feature pyramid denoted as $X$.
Each layer of features, $X_i$ where $i = 0, 1, 2, 3, 4$, possesses spatial sizes of $\frac{1}{2}$, $\frac{1}{4}$, $\frac{1}{8}$, $\frac{1}{16}$, and $\frac{1}{32}$ of the input image, respectively.
Subsequently, a Customized Feature Pyramid (CFP) is implemented, leveraging the extracted feature pyramid.
A lightweight Multi-Layer Perceptron (MLP) architecture is introduced to capture global long-range feature dependencies based on $X_4$, where the MLP layer takes the place of the multi-head self-attention module in a standard transformer encoder.
Compared to the transformer encoder utilizing multi-head attention, our proposed lightweight MLP architecture offers structural simplicity, reduced volume, and heightened computational efficiency (cf. Section III-B).
Furthermore, we employ a learnable visual center mechanism in conjunction with the lightweight MLP to aggregate the local corner regions of the input image.
This mechanism is denoted as the spatial EVC and is implemented on the top layer ($X_4$) of the feature pyramid. To enable efficient propagation of centralized visual information from the deepest feature to the shallow layer features, a Global Context Regulation (GCR) mechanism is proposed in a top-down fashion.
The explicit visual center information obtained from the deepest intra-layer feature regulates all frontal shallow features ($X_3$ to $X_1$) simultaneously. Finally, the aggregated features are directed into a decoupled head network for instance classification and bounding-box regression.



\subsubsection{Explicit Visual Center (EVC)}

\begin{figure*}[!htbp]
	\centering
	\includegraphics[width=1\linewidth]{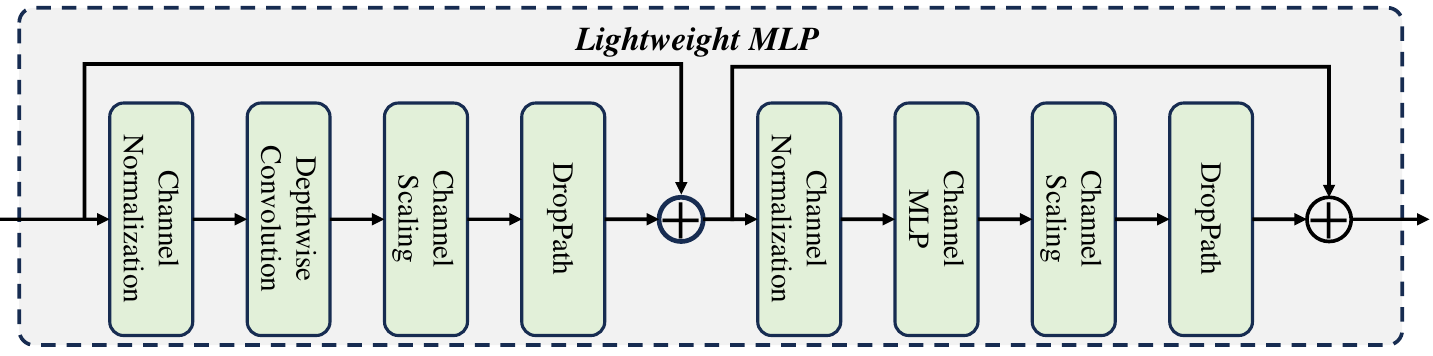}
	\caption{Illustration of lightweight MLP in the EVC module.}
	\label{MLP}
\end{figure*}

The Explicit Visual Center (EVC) framework proposed in this study comprises two parallel blocks designed to capture both global long-range dependencies and local corner regions within feature representations. The first block leverages a lightweight Multi-Layer Perceptron (MLP) to extract global information from the top-level features, denoted as $X_4$. In parallel, a learnable vision center mechanism is employed to aggregate intra-layer local region features, preserving the local corner regions. The resulting feature maps from these two blocks are concatenated along the channel dimension to form the output of the EVC for downstream recognition tasks.

Before mapping the original features directly, a Stem block is introduced between $X_4$ and the EVC to facilitate feature smoothing. The Stem block consists of a $7 \times 7$ convolutional layer with an output channel size of 256, followed by batch normalization and an activation function. The processes involved in the EVC can be formulated as follows:
\begin{equation}
X = \text{Cat}(\text{MLP}(X_{\text{in}}), \text{LVC}(X_{\text{in}}))
\end{equation}
where:
$\text{Cat}(\cdot)$ denotes feature map concatenation along the channel dimension,
$\text{MLP}(X_{\text{in}})$ and $\text{LVC}(X_{\text{in}})$ represent the output features of the lightweight MLP and the learnable visual center mechanism, respectively,
$X_{\text{in}}$ is the output of the Stem block, computed as:
\begin{equation}
X_{\text{in}} = \sigma(\text{BN}(\text{Conv}_{7 \times 7}(X_4)))
\end{equation}

The lightweight MLP within the EVC, as illustrated in Fig. \ref{MLP}, encompasses two key modules: a depthwise convolution-based module and a channel MLP-based block. The depthwise convolution-based module processes features output from the Stem module $X_{\text{in}}$ by first subjecting them to a depthwise convolution layer, followed by channel scaling and DropPath operations to enhance feature generalization and robustness. Meanwhile, the channel MLP-based module operates on the features obtained from the depthwise convolution-based module, also incorporating channel scaling and DropPath operations. The formulation for these processes is provided by:
\begin{equation}
\widetilde{X}_{\text{in}} = \text{DConv}(\text{GN}(X_{\text{in}})) + X_{\text{in}}
\end{equation}
where:
$\widetilde{X}_{\text{in}}$ represents the output of the depthwise convolution-based module,
$\text{DConv}(\cdot)$ denotes a depthwise convolution with a kernel size of $1 \times 1$, and
$\text{GN}(\cdot)$ denotes group normalization.

Similarly, the channel MLP-based module's operation can be expressed as:
\begin{equation}
MLP(X_{\text{in}}) = \text{CMLP}(\text{GN}(\widetilde{X}_{\text{in}})) + \widetilde{X}_{\text{in}}
\end{equation}
where:
$\text{CMLP}(\cdot)$ represents the channel MLP,
$\text{GN}(\cdot)$ denotes group normalization, and
$\widetilde{X}_{\text{in}}$ is the output of the depthwise convolution-based module.

\begin{figure*}[!htbp]
	\centering
	\includegraphics[width=1\linewidth]{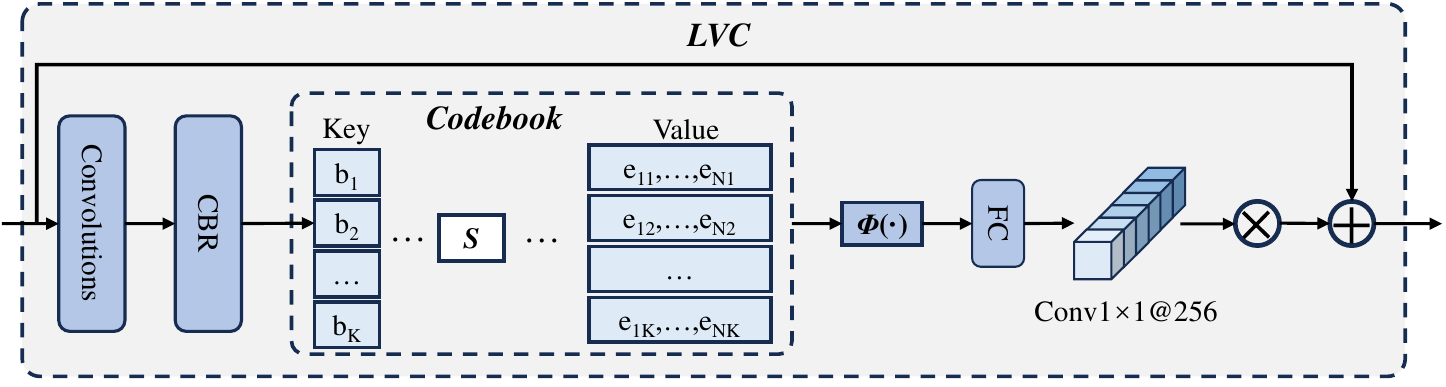}
	\caption{Illustration of LVC in the EVC module.}
	\label{LVC}
\end{figure*}

The learnable visual center (LVC) mechanism within the EVC, as illustrated in Fig. \ref{LVC}, is responsible for mapping the input feature $X_{\text{in}}$ to a set of $C$-dimensional features $X'_{\text{in}} = \{x'_1, x'_2, ..., x'_N\}$, where $N = H \times W$ represents the total number of input features. Subsequently, $X'_{\text{in}}$ learns an inherent codebook $B = \{b_1, b_2, ..., b_K\}$ containing $K$ codewords (visual centers) and a set of smoothing factors $S = \{s_1, s_2, ..., s_K\}$ for the visual centers. The key formulation for this process is:
\begin{equation}
e_k = \sum_{i=1}^{N} \frac{e^{-s_k \lVert x'_i - b_k \rVert^2}}{\sum_{j=1}^{K} e^{-s_j \lVert x'_i - b_j \rVert^2}} (x'_i - b_k)
\end{equation}
where:
$e_k$ represents the information of the whole image with respect to the $k$-th codeword,
$x'_i - b_k$ denotes the difference between the $N$ $C$-dimensional feature vectors and the $K$ codeword vectors,
$s_k$ is the smoothing factor of the $k$-th codeword, and
$\lVert\cdot\rVert$ denotes the L2 parametric operation.
The EVC also predicts a set of impact factors to highlight relevant categories, accomplishing this by mapping the feature $e$ to a $C \times 1 \times 1$ size using a fully connected layer, denoted as FC. The resulting feature representations are then obtained by performing channel-wise multiplication and addition operations between the input features $X_{\text{in}}$ and the impact factor coefficient.

\subsubsection{Global Centralized Regulation (GCR)}

The Global Centralized Regulation (GCR) technique proposed in this study addresses the computational efficiency of intra-layer feature regulation in a top-down manner within a feature pyramid. Recognizing that the deepest features contain the most abstract representations, the spatial EVC is initially implemented on the top layer ($X_4$) of the feature pyramid. Subsequently, the obtained features, inclusive of the spatial explicit visual centers, are utilized to regulate all lower-level features ($X_3$ to $X_1$) concurrently. This process involves upsampling the deep-layer features to the same spatial scale as the low-level features, concatenating them, and then downsampling by a $1 \times 1$ convolution to a channel size of 256.

The formulation for the computational efficiency improvement process is as follows:
\begin{equation}
Z = X_{in} \otimes (\delta(\text{FC}(e)))
\end{equation}
where:
$Z$ represents the local corner region features,
$\text{FC}$ denotes a fully connected layer,
$\delta(\cdot)$ represents the sigmoid function, and
$\otimes$ denotes channel-wise multiplication.

Finally, the GCR method concludes by performing a channel-wise addition operation between the features $X_{\text{in}}$ output from the Stem block and the local corner region features $Z$:
\begin{equation}
LVC(X_{\text{in}}) = X_{in} \oplus Z
\end{equation}
where $\oplus$ denotes the channel-wise addition operation.

Overall, the EVC and GCR modules collectively form a comprehensive framework for efficient feature extraction and regulation, encompassing the capture of long-range dependencies, preservation of local information, and top-down computational efficiency improvements within a feature pyramid for dense prediction tasks.

\section{Experiments}
\subsection{Datasets}
The benchmark FLIR \cite{FLIR2019, zhang2020multispectral} and LLVIP \cite{jia2021llvip} datasets are adopted in this work for evaluation.

The FLIR dataset consists of paired visible and infrared images for object detection, capturing both daytime and night scenes.
We utilized the FLIR-aligned dataset, as introduced by Zhang \emph{et al.}, which comprises 5142 RGB-IR image pairs, with 4129 images allocated for training and 1013 images for testing purposes.

The LLVIP dataset, developed by Jia \emph{et al.} in 2021, is a meticulously aligned visible and infrared object detection dataset tailored for low-light vision scenarios.
This dataset focuses on images captured in low-light environments, with a majority taken in very dark scenes.
The LLVIP dataset contains 15488 aligned RGB-IR image pairs, with 12025 images designated for training and 3463 images for testing.

\subsection{Evaluation Metrics}
We adopt mean Average Precision ($\text{mAP}$), $\text{mAP}_{50}$, and $\text{mAP}_{75}$ as the evaluation metrics.
The metrics necessitates the acquisition of ground-truth annotations and the corresponding predicted bounding boxes and class scores generated by the model for each image in the dataset.
Subsequently, the Average Precision (AP) is determined for each class by constructing precision-recall curves and quantifying the area under these curves.
For $\text{mAP}_{50}$, the assessment is confined to considering detections with an Intersection over Union (IoU) of 0.5 or higher as correct matches, whereas $\text{mAP}_{75}$ employs a more stringent IoU threshold of 0.75.
Following the generation of the AP values for each class, the mean of these values is computed to derive the $\text{mAP}$, $\text{mAP}_{50}$, and $\text{mAP}_{75}$ metrics.
The higher values of these metrics show better performance.

\subsection{Implemenration Details}
All experiments were executed on an NVIDIA GeForce RTX 3080 GPU.
During the training process, a batch size of 4 was employed, and the SGD optimizer was utilized with a momentum of 0.9 and a weight decay of $1\times 10^{-5}$.
The input image resolution was set to $640\times 640$.
All experiments were trained for 300 epochs with an initial learning rate of $1\times 10^{-2}$.
Furthermore, the data augmentation strategies used in YOLOv8 were applied to enhance input diversity.

\subsection{Ablation Study}

\begin{table}[tbp]
  \centering
  \caption{Ablation study on CRT-YOLO (in \%).}
  \setlength\tabcolsep{12pt} 
  \renewcommand\arraystretch{1.2}
  \begin{threeparttable}
  
  \begin{tabular}{c|cc|cc}
  \hline
  \multicolumn{1}{c|}{\multirow{2}{*}{Model}} & \multicolumn{2}{c|}{FLIR} & \multicolumn{2}{c}{LLVIP} \\
  \cline{2-5}
  & \makebox[0.04\textwidth][c]{mAP$_{50}$$\uparrow$} & \makebox[0.04\textwidth][c]{mAP$\uparrow$} & \makebox[0.04\textwidth][c]{mAP$_{50}$$\uparrow$} & \makebox[0.04\textwidth][c]{mAP$\uparrow$}\\
  \hline
  \textbf{CRT-YOLO} & \textbf{79.6} & \textbf{39.6} & \textbf{95.1} & \textbf{59.7} \\

  w/o EMA & 78.5 & 39.1 & 94.0 & 56.8 \\
  w/o EVC & 75.3 & 36.2 & 90.1 & 52.5 \\
  w/o MLP & 76.8 & 37.7 & 92.5 & 54.3 \\
  w/o LVC & 76.0 & 36.8 & 92.2 & 53.3 \\
  w/o GCR & 77.1 & 37.4 & 91.5 & 52.4 \\
  \hline
  \end{tabular}
  \end{threeparttable}
  \label{Ablation}
\end{table}

Table \ref{Ablation} presents the results of an ablation study conducted on the CRT-YOLO model, examining its performance under different conditions.
Each row in the table represents a different experimental condition, involving the removal of specific components or functionalities from the CRT-YOLO model.
The table includes metrics for mAP and mAP$_{50}$ on the FLIR and LLVIP datasets, serving as indicators of the model's performance in TIR object detection tasks.

The original configuration of the CRT-YOLO model demonstrated the highest performance, achieving 79.6\% mAP$_{50}$ and 39.6\% mAP on the FLIR dataset, as well as 95.1\% mAP$_{50}$ and 59.7\% mAP on the LLVIP dataset, serving as the baseline for subsequent ablation experiments.
Upon removing the EMA component, a slight performance decrease was observed, resulting in 78.5\% mAP$_{50}$ and 39.1\% mAP on the FLIR dataset, and 94.0\% mAP$_{50}$ and 56.8\% mAP on the LLVIP dataset, indicating a positive impact of EMA on the model's precision.
The absence of the EVC component led to a more substantial performance decline, with the model achieving 75.3\% mAP$_{50}$ and 36.2\% mAP on the FLIR dataset, and 90.1\% mAP$_{50}$ and 52.5\% mAP on the LLVIP dataset, underscoring the crucial role of EVC in enhancing precision on both datasets.
Furthermore, the absence of MLP and LVC, components of the EVC module, also contributed positively to the model's precision.
Upon removing the GCR component, the model experienced a decrease in performance, resulting in 77.1\% mAP$_{50}$ and 37.4\% mAP on the FLIR dataset, and 91.5\% mAP$_{50}$ and 52.4\% mAP on the LLVIP dataset, highlighting the significant role of GCR in enhancing precision, particularly in terms of mAP.
Overall, the results of the ablation study underscore the importance of each component or functionality within the CRT-YOLO model, providing valuable insights for optimizing the model's configuration to achieve the best performance in object detection tasks on both the FLIR and LLVIP datasets.

\subsection{Quantitative Results}

\begin{table}[htbp]
  \centering
  \caption{Comparison of the quantitative results on the FLIR and LLVIP dataset (in \%).}
  \setlength\tabcolsep{10pt} 
  \renewcommand\arraystretch{1.2}
  \begin{threeparttable}
  \begin{tabular}{l|cc|cc}

  \hline 
  \multicolumn{1}{c|}{\multirow{2}{*}{Model}} & \multicolumn{2}{c|}{FLIR} & \multicolumn{2}{c}{LLVIP} \\
  \cline{2-5}
  & \makebox[0.04\textwidth][c]{mAP$_{50}$$\uparrow$} & \makebox[0.04\textwidth][c]{mAP$\uparrow$} & \makebox[0.04\textwidth][c]{mAP$_{50}$$\uparrow$} & \makebox[0.04\textwidth][c]{mAP$\uparrow$}\\
  \hline 
  SSD \cite{liu2016ssd} & 65.5 & 29.6 & 90.2 & 53.5 \\
  RetinaNet \cite{lin2017focal} & 66.1 & 31.5 & 94.8 & 55.1 \\
  Cascade R-CNN \cite{cai2018cascade} & 71.0 & 34.7 & 95.0 & 56.8 \\
  Faster R-CNN \cite{ren2015faster} & 74.4 & 37.6 & 94.6 & 54.5 \\
  YOLOv3 \cite{redmon2018yolov3} & 73.6 & 36.8 & 89.7 & 52.8 \\
  YOLOv5 \cite{glenn_jocher_2021_5563715} & 73.9 & 39.5 & 94.6 & 61.9 \\
  YOLOF \cite{chen2021you} & 74.9 & 34.6 & 91.4 & 47.5 \\
  DDOD \cite{chen2021disentangle} & 72.7 & 33.9 & 94.3 & 56.6 \\
  DDQ-DETR \cite{zhang2023dense} & 73.9 & 37.1 & 92.1 & 56.6 \\ 
  YOLOv8 \cite{Jocher_Ultralytics_YOLO_2023}  & 76.3 & 36.7 & 93.9 & 58.6 \\
  \textbf{CRT-YOLO} & \textbf{79.6} & \textbf{39.6} & \textbf{95.1} & \textbf{59.7} \\
  \hline
  \end{tabular}
  \end{threeparttable}
  \label{Quantitative}
\end{table}

The comprehensive evaluation of the object detection models in Table \ref{Quantitative} reveals substantial variations in their performance across the FLIR and LLVIP datasets.
Notably, the CRT-YOLO model emerges as the top-performing model, exhibiting exceptional accuracy and robustness in object detection tasks involving both thermal and visible images.
Specifically, on the FLIR dataset, CRT-YOLO achieves an impressive mAP\textsubscript{50} of 79.6\%, indicative of its ability to accurately localize objects in thermal imagery, while also achieving a substantial mAP of 39.6\%.
This performance underscores the model's efficacy in handling the complexities and challenges associated with thermal imaging, including variations in temperature and environmental conditions.
Similarly, on the LLVIP dataset, CRT-YOLO demonstrates outstanding performance, attaining a remarkable mAP\textsubscript{50} of 95.1\% and an overall mAP of 59.7\%.
These results signify the model's excellence in accurately detecting and classifying objects in visible images, showcasing its adaptability and effectiveness across different imaging modalities.

In contrast, while other state-of-the-art object detection models such as SSD, RetinaNet, Faster R-CNN, YOLOv3, YOLOv5, YOLOF, DDQ-DETR, and YOLOv8 demonstrate competitive performance, they fall short of surpassing the comprehensive metrics achieved by CRT-YOLO.
For instance, the SSD model achieves 65.5\% mAP\textsubscript{50} and 29.6\% mAP on the FLIR dataset, and 90.2\% mAP\textsubscript{50} and 53.5\% mAP on the LLVIP dataset, showcasing strong performance but ultimately being outperformed by CRT-YOLO.
Similarly, other models, such as RetinaNet, demonstrate improvements over SSD, achieving 66.1\% mAP\textsubscript{50} and 31.5\% mAP on the FLIR dataset, and 94.8\% mAP\textsubscript{50} and 55.1\% mAP on the LLVIP dataset, yet still fall short of CRT-YOLO's performance.

Furthermore, models such as Faster R-CNN, YOLOv3, YOLOv5, YOLOF, DDQ-DETR, and YOLOv8 exhibit competitive performance, with mAP\textsubscript{50} ranging from 73.6\% to 76.3\% on the FLIR dataset and 89.7\% to 94.6\% on the LLVIP dataset, but none surpass the combined mAP\textsubscript{50} and mAP achieved by CRT-YOLO on both datasets.
This detailed analysis highlights the superior performance of CRT-YOLO in accurately localizing and classifying objects in thermal and visible images, positioning it as a leading solution for object detection tasks in diverse real-world scenarios.
Additionally, the impressive performance of CRT-YOLO underscores its potential for applications in critical domains such as surveillance, autonomous vehicles, and industrial automation, where accurate and robust object detection is paramount.

\subsection{Qualitative Results}

\begin{figure*}[!ht]
  \centering
  \includegraphics[width=1\linewidth]{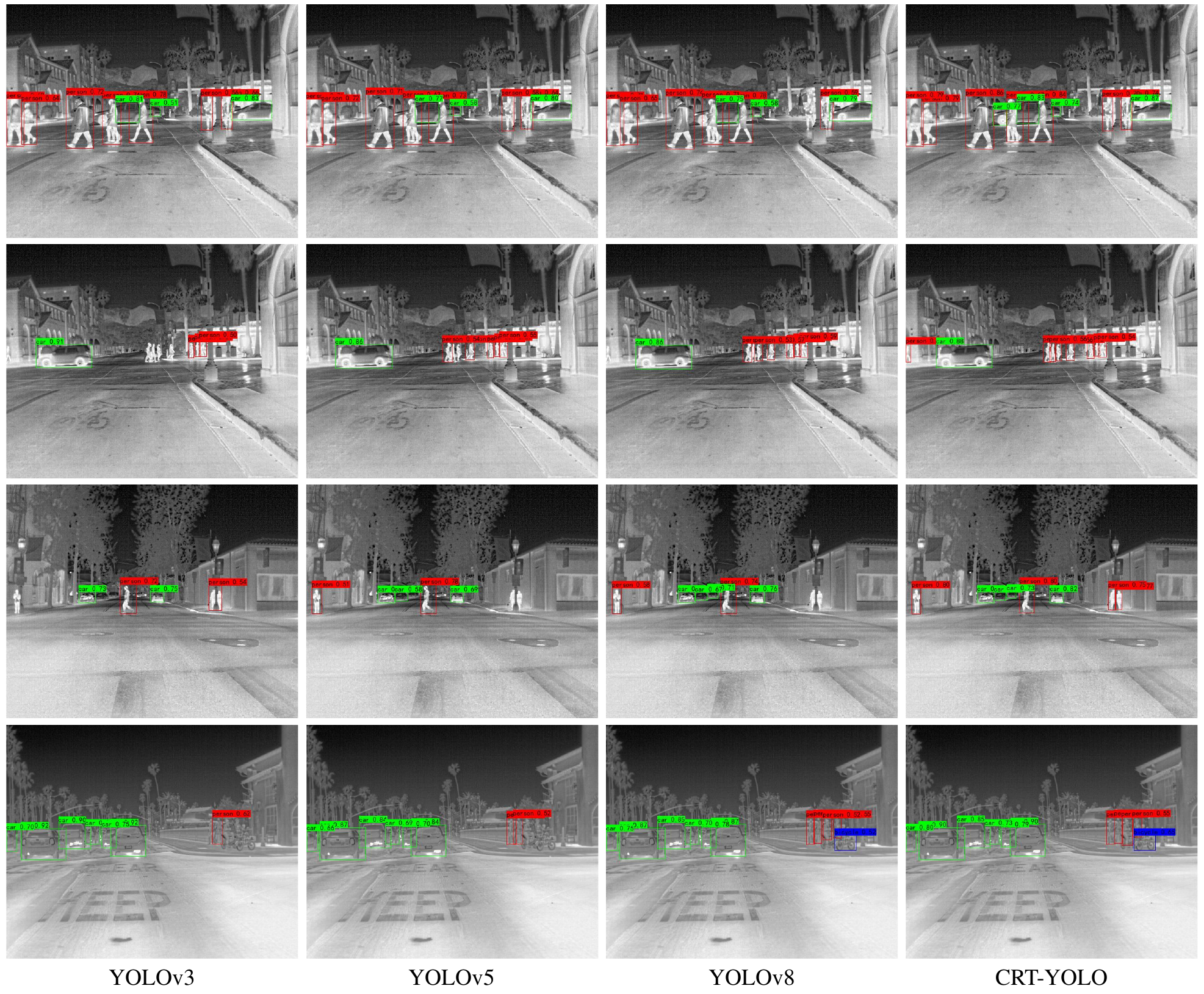}
  \caption{Qualitative comparisons between the baseline models and our CRT-YOLO on FLIR dataset.}
  \label{qualitative}
\end{figure*}


We show the qualitative results on FLIR dataset as it contains more categories, which are depicted in Fig. \ref{qualitative}.
The qualitative results illustrate the superior object detection capabilities of the proposed CRT-YOLO in TIR images when compared to the baseline methods, namely YOLOv3, YOLOv5, and YOLOv8.
Upon visual examination, it becomes apparent that CRT-YOLO consistently demonstrates its proficiency in both identifying and precisely localizing objects, thus resulting in significantly reduced occurrences of missed or misclassified objects.
In stark contrast, the baseline methods, particularly YOLOv3, YOLOv5, and YOLOv8, exhibit instances where objects are either inadequately detected or incorrectly labeled.
For instance, a notable observation from Fig. \ref{qualitative} is the consistent failure of the baseline methods to capture small-sized "person" instances in the second and third rows, whereas in the final row, both YOLOv3 and YOLOv5 are unable to detect "bicycles".
These specific examples serve as compelling evidence to emphasize the enhanced object detection capability of CRT-YOLO, underlining its potential for accurate and dependable object detection in challenging thermal imaging scenarios.

\section{Conclusion}
In conclusion, the proposed CRT-YOLO model has showcased exceptional performance and reliability in the domain of TIR object detection.
Through a comprehensive ablation study, it was evident that the individual components of CRT-YOLO, including the Efficient Multi-Scale Attention (EMA), Explicit Visual Center (EVC), and Global Centralized Regulation (GCR), each play a crucial role in enhancing the model's precision and overall performance.
The quantitative evaluation on the FLIR and LLVIP datasets further highlighted the superior accuracy and robustness of CRT-YOLO, surpassing the performance of other state-of-the-art object detection models.
Additionally, the qualitative comparison demonstrated CRT-YOLO's proficiency in accurately localizing and classifying objects in TIR images, showcasing its potential for real-world applications in surveillance, autonomous vehicles, and industrial automation.
Overall, the compelling results across the ablation study, quantitative evaluation, and qualitative comparison affirm the superior capabilities of CRT-YOLO, positioning it as a leading solution for accurate and reliable object detection in challenging thermal imaging scenarios.

\bibliographystyle{unsrt}
\bibliography{citation.bib}

\end{document}